\documentclass[10pt,twocolumn,letterpaper]{article}

\usepackage{cvpr}              

\usepackage{multirow}
\usepackage[dvipsnames]{xcolor}

\definecolor{cvprblue}{rgb}{0.21,0.49,0.74}
\usepackage[pagebackref,breaklinks,colorlinks,citecolor=cvprblue]{hyperref}
\usepackage{colortbl}
\usepackage{xcolor}
\usepackage{amssymb}
\usepackage{pifont}
\usepackage[nohyperlinks, nolist]{acronym}
\usepackage{siunitx}
\usepackage{cellspace, graphicx, makecell}
\usepackage{tikz}

\newcommand\rowincludegraphics[2][]{\raisebox{-0.45\height}{\includegraphics[#1]{#2}}}

\def\paperID{17} 
\def\confName{CVPR}
\def\confYear{2025}

\title{
    \AddToShipoutPicture*{\begin{tikzpicture}[overlay, remember picture]
        \node at ([yshift=-1.8cm]current page.north) {
            \normalsize\textcolor{gray}{This paper has been accepted for publication at the }
        };
        \node at ([yshift=-2.2cm]current page.north) {
            \normalsize\textcolor{gray}{IEEE/CVF Conference on Computer Vision and Pattern Recognition (CVPR), Workshops, Nashville, USA 2025}
        };
    \end{tikzpicture}}
    
    Towards Low-Latency Event-based Obstacle Avoidance on an FPGA-Drone
    
}

\author{
    Pietro Bonazzi$^{1}$, Christian Vogt$^{1}$,  Michael Jost$^{1}$ \\ Lyes Khacef$^{2}$, Federico Paredes-Vallés$^{2}$, Michele Magno$^{1}$ \\
    $^{1}$ETH Z\"urich $^{2}$ Sony Semiconductor Solutions Europe, Sony Europe B.V., Zürich, Switzerland \\ 
}

\begin{document}

\maketitle
\begin{abstract}

This work quantitatively evaluates the performance of event-based vision systems (EVS) against conventional RGB-based models for action prediction in collision avoidance on an FPGA accelerator. Our experiments demonstrate that the EVS model achieves a significantly higher effective frame rate (1 kHz) and lower temporal (-\qty{20}{\milli\second}) and spatial prediction errors (-\qty{20}{\milli\meter}) compared to the RGB-based model, particularly when tested on out-of-distribution data. The EVS model also exhibits superior robustness in selecting optimal evasion maneuvers. In particular, in distinguishing between movement and stationary states, it achieves a 59 percentage point advantage in precision (78\% vs. 19\%) and a substantially higher F1 score (0.73 vs. 0.06), highlighting the susceptibility of the RGB model to overfitting. Further analysis in different combinations of spatial classes confirms the consistent performance of the EVS model in both test data sets. Finally, we evaluated the system end-to-end and achieved a latency of approximately \qty{2.14}{\milli\second}, with event aggregation (\qty{1}{\milli\second}) and inference on the processing unit (\qty{0.94}{\milli\second}) accounting for the largest components. These results underscore the advantages of event-based vision for real-time collision avoidance and demonstrate its potential for deployment in resource-constrained environments.

\end{abstract}

\section*{Keywords}
Dataset, Obstacle Avoidance, Event-Based Cameras.

\section*{Funding}
This research was funded by the Swiss National Foundation (219943) and the Sony Research Award Program.

\section*{Reproducibility}

\begin{figure}[t!]
    \centering
    \begin{tabular}{c} 
       \includegraphics[width=0.96\columnwidth, trim=0cm 0cm 0cm 1.3cm, clip]{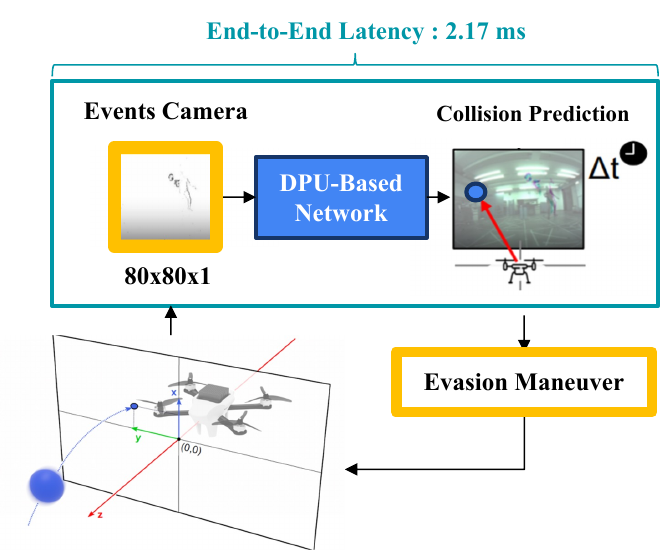} \\ 
    \end{tabular}
    \caption{\textbf{Overview of the proposed DPU-Based system for Obstacle Avoidance}. Our method achieves an end-to-end perception latency of sub 3ms.}
    \label{fig:intro}
\end{figure}

The data set and the code base can be installed following the instructions from the following link: \url{https://github.com/pbonazzi/eva}.
    
\begin{acronym}
\acro{SoC}[SoC]{System on Chip}
\acro{FPGA}[FPGA]{Field Programmable Gate Array}
\acro{EVS}[EVS]{Event Vision Sensor}
\acro{DPU}[DPU]{Deep Learning Unit}
\acro{DDR}[DDR]{Double Data Rate}
\end{acronym}
\section{Introduction}
\label{sec:intro}

Unmanned aerial vehicles (UAVs) are increasingly deployed across diverse applications, including logistics, surveillance, agriculture, and disaster response. As these autonomous systems operate in complex and dynamic environments \cite{wang2024dynamic}, ensuring real-time, reliable obstacle avoidance remains a critical challenge. Existing vision-based approaches predominantly rely on conventional RGB cameras due to their high spatial resolution and compatibility with established computer vision algorithms \cite{Xu_2023}. However, these frame-based sensors suffer from inherent limitations such as motion blur, latency, and reduced dynamic range—particularly in high-speed or low-light conditions \cite{yasin2020}.

Event-based vision sensors (EVS) offer an alternative paradigm by asynchronously capturing pixel-level intensity changes, enabling ultra-low latency, high temporal resolution, and enhanced robustness to lighting variations \cite{Gallego2022, gruel2023simultaneous, Bonazzi_2024_CVPR, Forrai_23_ICRA, monforte2023}. This sensing modality has demonstrated promising results in UAV applications, particularly for rapid motion detection and reactive navigation \cite{sanket2020evdodgenet, falanga2020, bisulco2021, paredes2024}. Recent advances in event-based vision have led to real-world UAV demonstrators \cite{doi:10.1126/scirobotics.adj8812}, but key challenges remain—namely, the need for robust, real-time decision-making in unpredictable environments and efficient onboard processing.

Despite the advantages of event-based vision, prior works have primarily focused on obstacle detection rather than direct action prediction for collision avoidance. Moreover, most implementations rely on software-based processing pipelines, which introduce latency and limit real-time responsiveness. Existing FPGA-based UAV architectures have demonstrated efficiency in general AI processing tasks \cite{tan2023, guo2021, kovari2021}, but few studies have explored FPGA acceleration for event-based action prediction in UAV navigation. The absence of comprehensive benchmarking between event-based and RGB-based methods on FPGA accelerators further limits our understanding of their comparative performance in real-world collision avoidance scenarios.

To bridge this gap, we propose an FPGA-event-based processing pipeline designed for real-time action prediction in UAV collision avoidance. Unlike prior works that primarily focus on obstacle detection or trajectory estimation, our approach evaluates event-based and RGB modalities to predict both the position and timing of potential collisions with high precision. 

We quantitatively evaluated event-based vision systems against conventional RGB-based models for collision avoidance, demonstrating superior temporal resolution and spatial accuracy. Additionally, we provide a comprehensive analysis of event-based action prediction across two test scenarios, including out-of-distribution trajectories, and release all 238 recordings as an open dataset to foster further research. To ensure practical feasibility, we implement an FPGA-based processing pipeline that achieves a low-latency inference time of approximately \qty{2.14}{\milli\second}, enabling real-time UAV navigation which features an efficient frame aggregation that processes raw \ac{EVS} events, accumulating them into \(80 \times 80 \) pixel frames with fully configurable accumulation parameters. A visual overview of the proposed system is shown in Fig.~\ref{fig:intro}. 
\section{Related Work}

Recent advancements in UAV technology have increasingly focused on integrating FPGA-based architectures to enhance real-time processing capabilities. This section reviews notable contributions in this domain, highlighting existing research and identifying key differences from our approach.

\subsection{Event-Based Obstacle Avoidance}

Prior research has explored event-based vision systems for UAV obstacle avoidance using both traditional and learning-based approaches. Falanga et al. \cite{falanga2020} and Rodriguez-Gomez et al. \cite{rodriguez2022} leveraged optical flow and clustering techniques for dynamic obstacle avoidance in quadrotors and ornithopter robots, respectively. Falanga \cite{falanga2019} presented a theoretical framework that jointly considers perception latency and actuation limitations to analyze the maximum speed at which a robot can safely navigate through unknown, cluttered environments, highlighting the trade-offs between sensing latency, sensing range, and actuation capabilities for high-speed navigation. More recent works have shifted towards data-driven solutions: Sanket et al. \cite{sanket2020evdodgenet} and Bhattacharya et al. \cite{bhattacharya2024monocular} developed monocular event-camera-based static obstacle avoidance methods, while Bisulco et al. \cite{bisulco2021} introduced spatio-temporal neural networks for fast-motion perception. Additionally, Paredes-Valles et al. \cite{paredes2024} trained a spiking neural network to enable autonomous flight using raw event-based camera data.

While these works focus on obstacle detection and navigation, our study extends the scope by evaluating action prediction in a collision avoidance scenario. Unlike prior event-based models, i.e. \cite{falanga2019}, our approach emphasizes real-time decision-making robustness on an FPGA accelerator. Furthermore, we introduce a lightweight inference pipeline and conduct extensive comparisons with RGB-based models, highlighting the advantages of event-based processing in terms of accuracy and robustness.

\subsection{FPGA-Based Architectures for UAVs}

Several studies have explored FPGA-based UAV architectures to improve computational efficiency and response time. Guo et al. \cite{guo2021} proposed a hybrid UAV architecture integrating FPGA, ARM, and GPU components to overcome hardware constraints and enhance real-time image processing. Similarly, Kovari and Ebeid \cite{kovari2021} introduced MPDrone, an FPGA-based system designed for intelligent real-time drone operations, offloading AI computations to the FPGA while maintaining control processes on a CPU. Tan et al. \cite{tan2023} focused on FPGA-based acceleration for simultaneous localization and mapping (SLAM), demonstrating efficiency improvements in SLAM processing.

In contrast, our work targets action prediction for collision avoidance rather than general UAV computation. We demonstrate that event-based models can outperform RGB-based methods on FPGA accelerators in both speed and accuracy, achieving real-time inference with minimal latency (\qty{2.14}{\milli\second} vs estimated \qty{21.14}{\milli\second}). Our approach prioritizes not only computational efficiency but also decision-making reliability in out-of-distribution scenarios.

\subsection{Open-Source Frameworks and Platforms}

Nyboe et al. \cite{Nyboe_2022} presented MPSoC4Drones, an open framework integrating ROS2, PX4, and FPGA, facilitating the development of advanced UAV applications. Foehn et al. \cite{Foehn_2022} introduced Agilicious, an open-source and open-hardware agile quadrotor designed for vision-based flight, promoting accessibility and collaboration in UAV research. SwiftEagle  \cite{vogt2024IROS} is an open-source FPGA-based UAS platform integrating dual RGB/EVS cameras with a multi-sensor subsystem, achieving sub-microsecond time resolution and low-latency EVS event frame rendering.

Our work complements these efforts by introducing a publicly available dataset and an efficient event-based inference pipeline for FPGA-based action prediction. Unlike existing platforms that primarily support general perception tasks, we provide an end-to-end evaluation tailored for collision avoidance, bridging the gap between perception and real-time decision-making in UAV applications.

\begin{figure*}[ht!]
    \centering
    \caption{\textbf{Position of the ball relative to the drone across all recordings of our dataset.} The 3D position of the drone and the ball together with the respective timestamps are used to obtain the ground truth place and time of collision of the ball.}
    \includegraphics[width=0.98\linewidth]{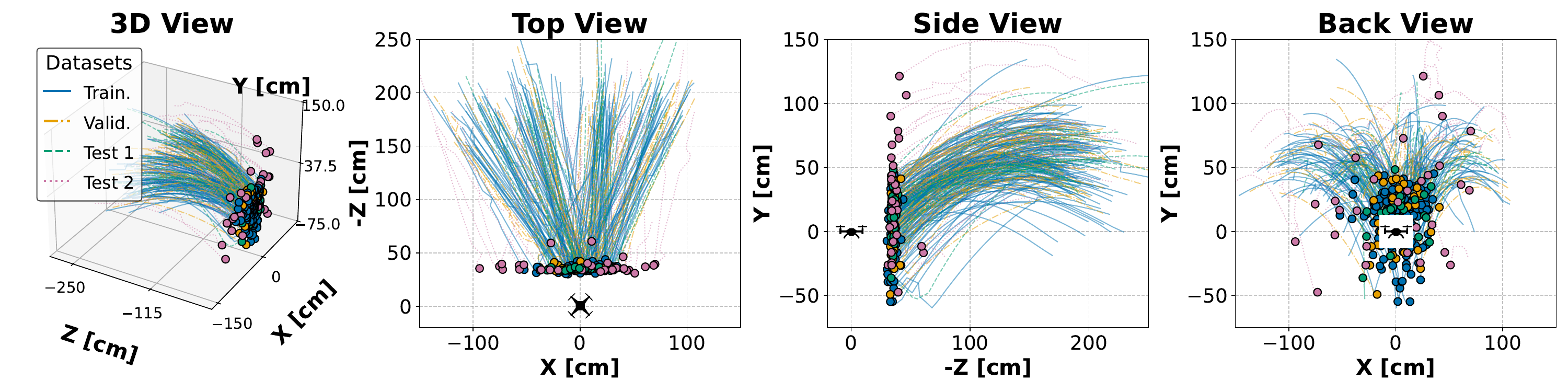}
    \label{tab:dataset_trajectories}
\end{figure*}

\section{Methodology}
\label{sec:methodology}

\subsection{Data Collection}

This study utilizes the SwiftEagle platform \cite{vogt2024IROS}, an open-source, FPGA-based unmanned aerial vehicle (UAV) designed for low-latency sensing and real-time decision-making. The system integrates a frame-based camera (SONY IMX219) and an event-based vision sensor (Prophesee GenX320). At the core of this platform is a custom FPGA design that facilitates dual-camera interfaces, supporting both RGB and event-based vision sensing (EVS). This combination allows for high-speed, low-latency data acquisition, which is crucial for motion prediction and real-time analysis of fast-moving objects, such as the ball featured in our dataset. The ball used is a Betzold sport softball, with an \qty{15}{\centi\meter} diameter and made from blue polyurethane foam (weight: \qty{65}{\gram}). The internal foam was removed and replaced with six infrared LEDs, which were powered by batteries. A plexiglass barrier was placed between the ball and the drone to protect the sensors from collisions, though this introduced reflections and vibrations that degraded the visual data quality.

The dataset consists of 238 recordings of ball throws. For training purposes, the network was only exposed to the last 45 timestamps of each recording, prior to the ball's impact with the plexiglass barrier.

This dataset serves as a comprehensive benchmark intended to extend NeuroBench \cite{yik2025neurobenchframeworkbenchmarkingneuromorphic} and explore the dynamics of moving objects in 3D environments. It includes multi-modal recordings of ball trajectories under a variety of conditions, such as elevation angles ranging from \qty{0.03}{\meter} to \qty{1.99}{\meter}, flight distances between \qty{0.51}{\meter} and \qty{2.28}{\meter}, and velocities varying from \qty{0.07}{\meter\per\second} to \qty{10.87}{\meter\per\second}. These diverse conditions provide a robust foundation for analyzing motion prediction, collision dynamics, and sensor fusion techniques.

The experiments were conducted in a controlled setting utilizing a Vicon motion capture system. The setup involved a square arena surrounded by multiple cameras to precisely capture the 3D trajectories of both the ball and the drone. Reflective markers were placed on the ball and on the drone to enable accurate tracking of their 3D positions.

To illustrate the multi-modal nature of the dataset, Table~\ref{fig:images_over_time} presents a sample recording with annotations detailing the timing and location of ball collisions. The rows in this table correspond to the various data modalities captured, including RGB camera footage, event camera data, and 3D positional data for both the drone and the ball.

\begin{table}[ht]
\centering 
\begin{tabular}{m{0.5cm} |ccc}
\toprule
TTC & RGB Camera & Event Camera & 3D Positions\\
\midrule
\centering \rotatebox{90}{400ms} & \multicolumn{3}{c}{\rowincludegraphics[trim={130 0 60 50}, clip, width=0.8\linewidth]{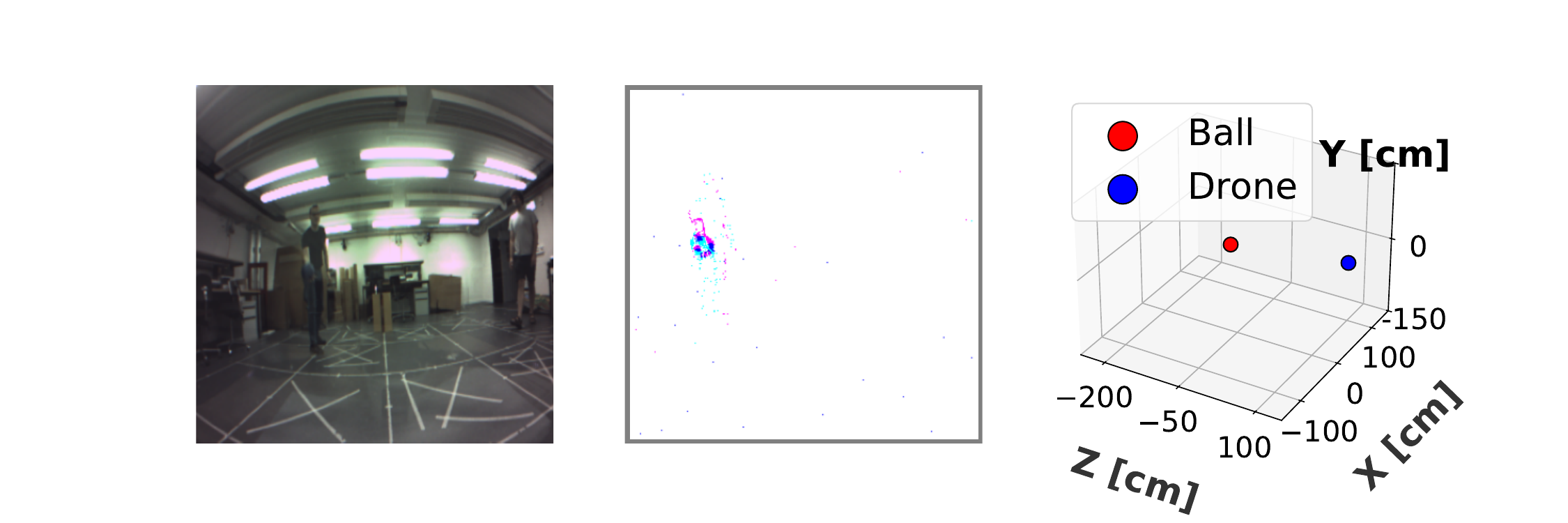}} \\
\centering \rotatebox{90}{200ms} & \multicolumn{3}{c}{\rowincludegraphics[trim={130 0 60 50}, clip, width=0.8\linewidth]{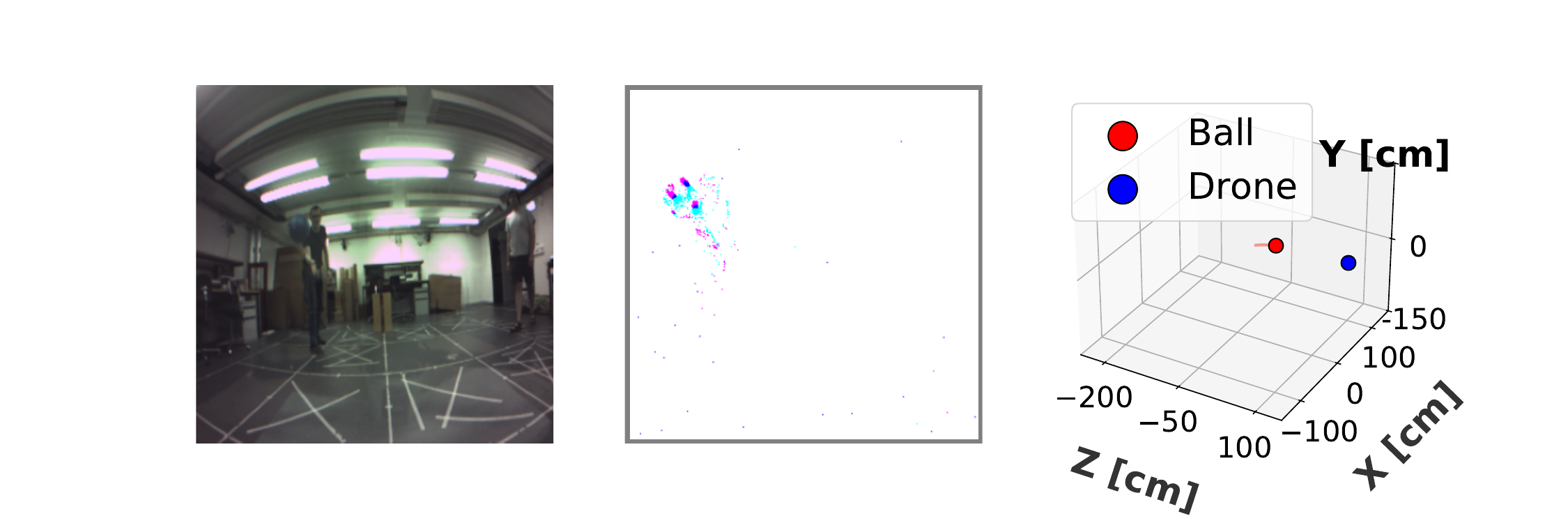}} \\
\centering \rotatebox{90}{001ms} & \multicolumn{3}{c}{\rowincludegraphics[trim={130 0 60 50}, clip, width=0.8\linewidth]{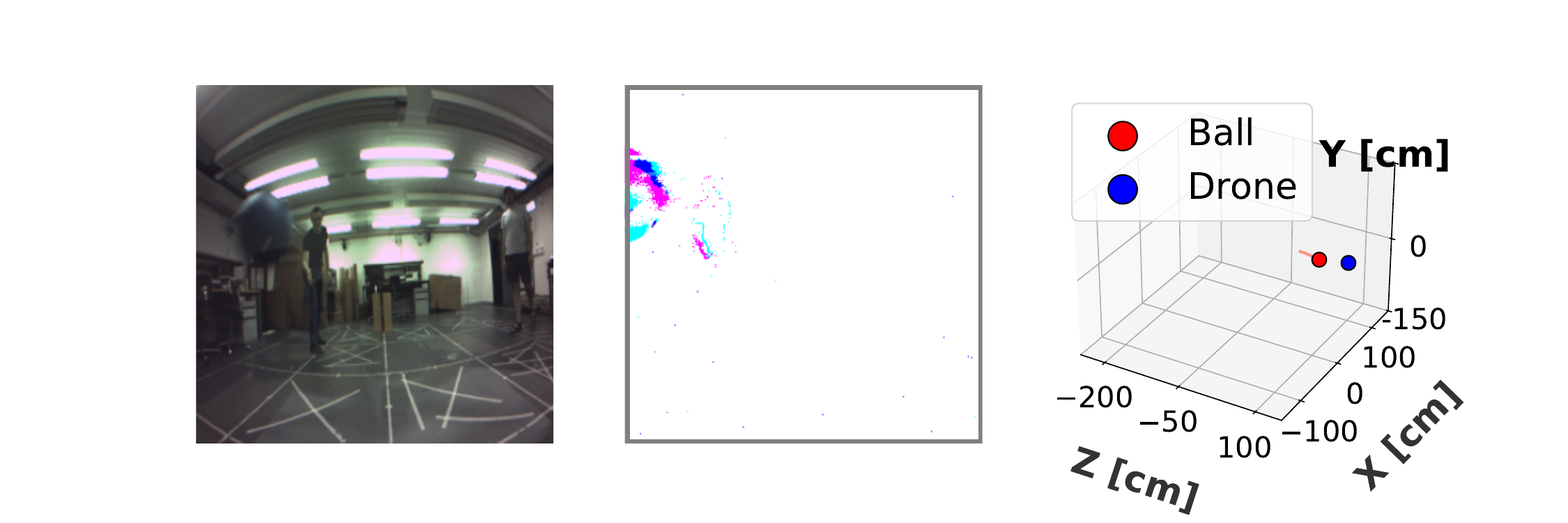}}\\
\bottomrule
\end{tabular}
\caption{\textbf{Visualization of a sample recording.} The 3D position of the ball and the drone are used to compute the time to collision (TTC).}
\label{fig:images_over_time}
\end{table}

For an overview of the entire dataset, Figure ~\ref{tab:dataset_trajectories} shows the ball trajectories across all recordings. The dataset is divided into distinct subsets for training, validation, and testing, with different colors representing these groups. Additionally, a drone icon provides spatial reference. The dataset was split by first identifying the collision points and then dividing the unique recording IDs into training (60\%), validation (20\%), and test (20\%) sets, ensuring a balanced representation of collision locations.

\begin{table}[ht]
\centering
\begin{tabular}{c|c|c|c|c|c}
\toprule
\multirow{3}{*}{\textbf{Data Splits}} & \multicolumn{4}{c|}{\textbf{Place of Collision}} & \textbf{Total} \\ \cmidrule{2-5}
& \multicolumn{2}{c|}{\textbf{Up}}  & \multicolumn{2}{c|}{\textbf{Down}} & \textbf{\# of} \\
& \textbf{Left} & \textbf{Right} & \textbf{Left} & \textbf{Right} & \textbf{samples} \\
\midrule
\rowcolor{blue!10} \textbf{Train.} & 22 & 35 & 38 & 47 & 142 \\
\rowcolor{orange!10} \textbf{Valid.} & 9 & 9 & 14 & 8 & 40 \\
\rowcolor{green!10} \textbf{Test 1} & 4 & 5 & 10 & 7 & 26 \\
\rowcolor{red!10} \textbf{Test 2} & 6 & 5 & 10 & 9 & 30 \\ \midrule
\textbf{Total} & 41 & 54 & 72 & 71 & 238 \\
\bottomrule
\end{tabular}
\caption{\textbf{Distribution of collision points across datasets.} The dataset is split ensuring a balanced representation of collision locations.}
\label{tab:collision_statistics}
\end{table}

In this study, we also provide a second test dataset (Test 2), which features a significantly more diverse set of collisions. This test dataset was recorded in a separate date with respect to the first test set and was kept separate to compare the robustness of EVS and RGB-based models. 

Table~\ref{tab:collision_statistics} displays the distribution of collision points across different datasets.

\subsection{Neural Network}

To predict the trajectory and time to collision of an object using event-based and RGB data, we designed a neural network model tailored for efficient deployment on FPGA hardware. This section details the key aspects of the model, including its input representation, architectural design, training procedures and evaluation methodology.

\subsubsection{Input Representation}

The input representation of the event camera was designed to facilitate efficient memory management and computational deployment on FPGA hardware.

For the event-based input representation, event streams are processed using a fixed time window of duration \(T\) (\(T = \qty{20}{\milli\second}\) ms unless otherwise noted). Within this window, events are accumulated on a per-pixel basis, separately for both positive and negative polarities. The final event representation is obtained by computing the pixel-wise difference between the two polarities, resulting in a tensor of dimensions \(80 \times 80 \times 1\).

For the RGB input representation, raw frames (\(T = \qty{20}{\milli\second}\)) are captured and resized to a fixed resolution of \(80 \times 80 \times 3\). Similar to event data, no temporal multi-dimension is applied to the RGB input unless explicitly stated.

Each input image and event frame is normalized to the range [-1, 1] at every iteration to be compatible with the input required by the DPU of the FPGA. This normalization influenced the choice of the activation function, which is a leaky ReLU \cite{xu2015empiricalevaluationrectifiedactivations}, to handle the presence of negative data in the input. Furthermore, the output data was normalized using min-max normalization, with ranges corresponding to the splits of the train dataset. In contrast, test 2 exhibited significantly different collision points, leading to different ranges for its normalization.

\subsubsection{Architecture}

The proposed model incorporates an encoder structure which is employed for both event-based and RGB image inputs through separate networks.

It features 6 convolutional layers (\(3 \times 3\) kernel size, stride of 1) followed by leaky ReLU \cite{xu2015empiricalevaluationrectifiedactivations} activation functions, designed to learn spatial hierarchies from the input data progressively.

Finally, a fully connected layer transforms the output of each encoders separately into a 3-dimensional prediction: the \(x\) and \(y\) coordinates of the collision relative to the drone, and \(t\), the estimated time to collision. In this last layer, no activation function is used.


The model has a total of 192M floating-point operations per forward pass and 278 thousands trainable parameters. When operating at FP32 precision, the model size is approximately 1.1 MB. For INT8 precision, the size is reduced to about 278 KB.

\subsubsection{Loss Function and Training Details}

The task involves predicting the relative XY position of the collision point of a ball of fixed size with respect to the drone's location (static across all sequences) and the time to collision. For this purpose, the loss function used is the Mean Squared Error (MSE) for both predictions. The 3D positions of the drone and the ball are obtained from a Vicon motion capture system. The collision point is computed as the minimum distance between the two 3D trajectories (the ball and the drone) over the recording period.

We used a batch size of 32 and an initial learning rate of 1e-4. The training process incorporated stochastic weight averaging, gradient clipping, and early stopping with a patience of 10 epochs to prevent overfitting. Optimization was performed using the Adam optimizer \cite{kingma2017adammethodstochasticoptimization} with a weight decay of 1e-5. Additionally, a learning rate scheduler was employed to reduce the learning rate exponentionally with a $\gamma$ of 0.98. Dropout regularization is employed at each stage to enhance generalization.

\subsubsection{Evaluation Method}

The evaluation of the model is based on six distinct classes: left, right, up, down, move, and stay. The sign of the spatial prediction is used to assign the ball's trajectory to the appropriate quadrant (left, right, up, or down), based on the relative position of the collision with respect to the drone. The 'move' and 'stay' classes are determined by comparing the time to collision with a threshold of \qty{100}{\milli\second}: if the time to collision is lower than \qty{100}{\milli\second}, the ball is classified as 'move'; otherwise, it is classified as 'stay'. The models are evaluated using accuracy and F1 score for each class. F1 score is used to balance precision and recall, particularly useful when dealing with imbalanced datasets or when the cost of false positives and false negatives differs. 

In the context of the 'move' and 'stay' classes, false negatives (where a move is classified as stay) are a primary concern since these actions determining whether the ball is moving quickly enough to warrant attention. At the same time, for the spatial classes (left, right, up, down), accurately predicting the place of collision is equally crucial. An incorrect classification of the collision location could result in significant errors in decision-making.

The model's evaluation is conducted at two levels: "window-level" and "launch-level". Window-level predictions classify the ball's trajectory and movement status for each time window independently. Launch-level predictions aggregate these window-level outputs using majority voting to determine the final trajectory classification for the entire recording. Temporal classes (move/stay) are only evaluated at the window level.

\subsection{Hardware Implementation}
\begin{figure}[ht]
    \centering
    \includegraphics[width=0.5\textwidth]{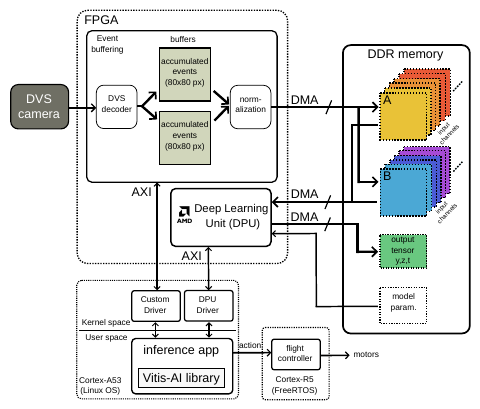}
    \caption{\textbf{Hardware/Software Co-design approach} integrating the detection network on the SwiftEagle platform}
    \label{fig:hardware_design}
\end{figure}

We integrate the presented prediction framework on the SwiftEagle's \cite{vogt2024IROS} onboard \ac{SoC} by a hardware-software co-design approach, depicted in \Cref{fig:hardware_design}. 

On the programmable hardware realized with the \ac{FPGA} fabric, both, the custom optimized \ac{EVS} frame aggregation, as well as the commercial inference module (AMD Xilinx \ac{DPU}) are located.

The software part is split into two main components: orchestration of the inference and reaction to the inference results. The orchestration is handled by a custom application for the on-board quad-core ARM Cortex A53 processors running Yocto \footnote{\url{https://github.com/Xilinx/meta-xilinx}} Linux (APU). Finally, evaluation of the inference results and taking appropriate action is handled by the on-board real time processing unit (RPU) ARM Cortex-R5, which also runs the SwiftEagle flight stack.

The frame aggregation on the programmable logic directly processes raw \ac{EVS} events, accumulating them into \(80 \times 80 \)  pixel frames with fully configurable accumulation parameters. The proposed FPGA-acquisition block can be configured either based on accumulation time or by the number of events. 

Aggregated \ac{EVS} frames are then normalized and transferred to the \ac{DDR} memory upon completion. To prevent loss of events during the data transfer from fabric to \ac{DDR}, we implemented a double-buffering scheme on the fabric: while one buffer is transferred to DDR, the other buffer is used for aggregation, and vice versa. The  \ac{DPU} is configured as such: type \emph{dpuczdx8g}, with architecture \emph{1600} and 1 core. The \ac{DPU} can directly read the accumulated frames from DDR memory and also stores the inference outputs again in \ac{DDR} memory. 

On the software side, the custom developed application receives a hardware trigger when a new \ac{EVS} frame is available in the \ac{DDR} memory and reacts by triggering the \ac{DPU} to run an inference. The application also ensures setup of the \ac{DPU} with the model, as well as assigning correct memory addresses for frames and inference results. This approach allows the usage of the tool-chain provided by the \ac{DPU} manufacturer AMD. The Firmware of the on-board microcontroller finally implements the decision logic for evasion decision.

\section{Experiments}
\label{sec:experiments}

In this section, we present a comprehensive evaluation of the proposed models, focusing on their performance across multiple aspects relevant to real-world deployment. We compare RGB-based and event-based vision networks under different conditions, analyzing their precision at varying frame rates, robustness across different action classes, and effectiveness in predicting complete motion sequences. Additionally, we assess the models' trajectory estimation capabilities and investigate the overall system latency to ensure suitability for real-time applications.

\subsection{Comparison of RGB and EVS models}

To provide a thorough assessment of our approach, we begin by comparing the performance of RGB-based and event-based vision models across different scenarios. This comparison is crucial for understanding the advantages of event-based sensing in terms of temporal resolution, robustness, and computational efficiency.

\subsubsection{Average precision at different frame rates}

To evaluate the perception latency limits of RGB and event-based models in different deployment scenarios, we analyzed their performance across a range of frame rates.

\begin{figure}[ht]
    \centering
    \begin{tabular}{c}
        \includegraphics[width=0.47\textwidth]{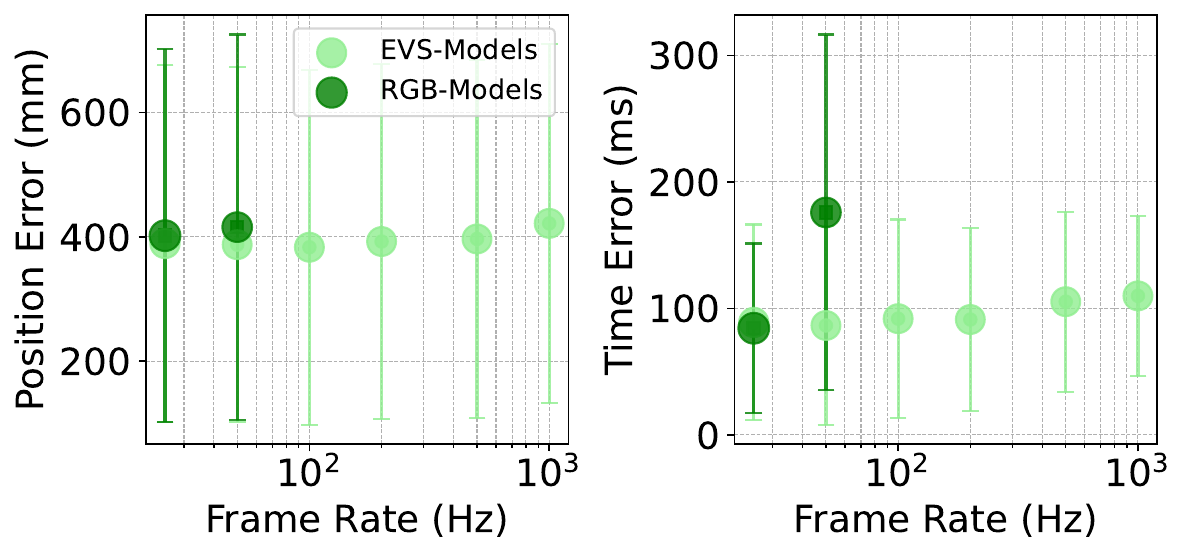}  \\
    \end{tabular}
    \caption{\textbf{Performance of RGB and event-based models at different frame rates.} On average, EVS-only models can increase throughput and outperform RGB models running at similar frame rates.}
    \label{fig:frame_rate_over_error}
\end{figure}

For the event-based model, we define the frame rate as $1/T$, where $T$ represents the accumulation time window. In contrast, the RGB camera operates at a fixed frame rate of 50 frames per second \cite{vogt2024IROS}. To enable a fair comparison at lower frame rates, we artificially downsampled the RGB input by concatenating consecutive frames. This aggregation improved both temporal and spatial performance, as shown in Fig.~\ref{fig:frame_rate_over_error}. Notably, the reduction in time error suggests that a single RGB frame lacks sufficient information for precise temporal prediction.

A frame rate of 25Hz—corresponding to the first points from the left in both plots of Fig.~\ref{fig:frame_rate_over_error}—was achieved by applying two consecutive, non-overlapping \qty{20}{\milli\second} windows in the EVS models.

The results further highlight that EVS-based models consistently outperform their RGB counterparts, exhibiting lower errors even at higher frame rates. Crucially, EVS models can operate at significantly higher frame rates while still maintaining lower position error than RGB models running at 50Hz.

Interestingly, beyond a certain frame rate, position-of-collision accuracy remains relatively stable, suggesting that spatial information within a \qty{1}{\milli\second} event accumulation window may already be sufficient to solve the task effectively.

\subsubsection{Single classes evaluation: EVS vs RGB}

A critical aspect of deploying AI models in real-world scenarios is their ability to generalize across varying conditions. This section compares the robustness of event-based and RGB models in action classification, particularly focusing on their robustness to unseen data and computational efficiency.

\begin{table}
\centering
\resizebox{\linewidth}{!}{%
\begin{tabular}{c | c | cc | cc | cc}
\toprule
\multirow{2}{*}{\textbf{Data}} & \multirow{2}{*}{\textbf{Model}} & \multicolumn{2}{c|}{\textbf{Left/Right}} & \multicolumn{2}{c|}{\textbf{Up/Down}} & \multicolumn{2}{c}{\textbf{Move/Stay}}  \\ 
 &  & \textbf{Acc} & \textbf{F1} & \textbf{Acc} & \textbf{F1} & \textbf{Acc} & \textbf{F1}  \\ 
\midrule

Test & RGB & 63 & 0.6 & 69 & 0.7 & 89 & 0.87 \\   \cmidrule{2-8}
1 & EVS & 68 & 0.66 & 64 & 0.65 & 86 & 0.85 \\   \midrule
Test & RGB & 58 & 0.51 & 50 & 0.51 & 19 & 0.06 \\   \cmidrule{2-8}
2 & EVS & 69 & 0.69 & 63 & 0.63 & 74 & 0.7 \\   \bottomrule
\end{tabular}%
}
\caption{\textbf{Comparison of collision prediction models.} Individual window-level class evaluations for two test sets.}
\label{tab:rgb_vs_evs_results}
\end{table}

Both models were evaluated at the RGB frame rate to ensure a fair comparison. The EVS model used a \qty{20}{\milli\second} accumulation window with a single event bin, making it slightly more lightweight than the RGB model, which relied on three input channels. This difference translated into a lower computational cost for EVS (192 Mflops) compared to RGB (202 Mflops).

As shown in Table \ref{tab:rgb_vs_evs_results}, the results indicate that the EVS model generalizes better, exhibiting less overfitting. In the first test set, both models performed similarly, with RGB slightly ahead in Move/Stay (89\% vs. 86\%). However, in the second test set, RGB's accuracy in Move/Stay dropped sharply (89\% to 19\%), whereas EVS remained stable (86\% to 78\%). The same trend is seen in F1 scores: RGB fell from 0.87 to 0.06, while EVS maintained a higher 0.84 to 0.73. These findings highlight EVS as a more robust and efficient alternative to RGB, particularly for real-world deployment where test conditions vary.

\subsection{EVS-based action evaluation}

Beyond isolated comparisons of RGB and EVS models, it is essential to evaluate how well event-based models capture entire motion sequences over time. In this section, we focus on the ability of the EVS model to maintain consistency in action predictions across different test datasets. We first assess performance at both window and recording levels to determine how well the model tracks motion. Then, we analyze its trajectory estimation and time-to-collision predictions, providing deeper insights into its reliability for real-world deployment.

\subsubsection{Window and Recording-Level Analysis}

\begin{table*}[h]
\centering
\begin{tabular}{cc|c|c|c|c|c|c|c|c}
\toprule
 && \multicolumn{4}{c|}{\textbf{Test Dataset 1}} & \multicolumn{4}{c}{\textbf{Test Dataset 2}} \\ \midrule
 \multicolumn{2}{c|}{\textbf{Spatial Classes}} & \multicolumn{2}{c|}{\textbf{Window-level}} &  \multicolumn{2}{c|}{\textbf{Launch-level}}  & \multicolumn{2}{c|}{\textbf{Window-level}} &  \multicolumn{2}{c}{\textbf{Launch-level}}  \\ \midrule
 \textbf{Horizontal} & \textbf{Vertical} &  \textbf{Acc} & \textbf{F1} &  \textbf{Acc} & \textbf{F1} &  \textbf{Acc} & \textbf{F1} &  \textbf{Acc} & \textbf{F1}  \\ \midrule

Left & Up & 0.77 & 0.42 & 0.77 & 0.50 & 0.76 & 0.32 & 0.77 & 0.36 \\ \midrule
Right & Up & 0.79 & 0.45 & 0.85 & 0.60  & 0.78 & 0.35 & 0.83 & 0.44  \\ \midrule
Left & Down & 0.61 & 0.50 & 0.62 & 0.50 & 0.65 & 0.43 & 0.63 & 0.42  \\ \midrule
Right & Down & 0.70 & 0.32 & 0.77 & 0.40  & 0.65 & 0.49 & 0.70 & 0.57  \\ \midrule \midrule 

\multicolumn{2}{c|}{\textbf{Temporal Classes}} & \multicolumn{4}{c|}{\textbf{Window-level}}  & \multicolumn{4}{c}{\textbf{Window-level}}  \\ \midrule
\multicolumn{2}{c|}{\textbf{Type}} & \textbf{Acc} & \textbf{F1} & \textbf{Recall} & \textbf{Precision} & \textbf{Acc} & \textbf{F1} & \textbf{Recall} & \textbf{Precision}\\ \midrule
  
\multicolumn{2}{c|}{Move} &  \multirow{2}{*}{0.86} & 0.92 & 0.97 & 0.87 & \multirow{2}{*}{0.78} & 0.87 & 0.94 & 0.81 \\
\multicolumn{2}{c|}{Stay} &  & 0.49 & 0.36 & 0.73 & & 0.13 & 0.09 & 0.27 \\ \bottomrule 

\end{tabular}
\caption{\textbf{Performance metrics at window-level and recording-level on the two test set.} All recordings have one collision, therefore at launch-level only spatial predictions with majority voting are presented.}
\label{tab:window_results}
\end{table*}

The spatial analysis in Table~\ref{tab:window_results} reveals that aggregating predictions via majority voting at the launch-level generally enhances performance relative to window-level metrics. For example, the “Right Up” class in Test Dataset 1 shows an increase in the F1 score from 0.45 to 0.60 when moving from window level to launch level evaluation, indicating that temporal aggregation helps mitigate transient misclassifications. Performance also varies by directional combination, with “Right-Up” achieving consistently higher accuracy and F1 scores compared to other classes. Moreover, Test Dataset 2 tends to yield slightly lower F1 scores than Test Dataset 1, suggesting that variations in dataset characteristics or noise levels impact spatial prediction accuracy.

Temporal analysis indicates that the model performs robustly for the “Move” class, achieving high accuracy, recall, and precision across both datasets. In contrast, predictions for the “Stay” class are markedly weaker, as evidenced by low F1 scores and recall, particularly in Test Dataset 2. This discrepancy implies that the model struggles to reliably identify stationary periods, which may be attributable to inherent class imbalances or the subtlety of motion cessation cues.

Overall, the benefits of temporal aggregation and the sensitivity to dataset variability should be considered in future work to enhance the model’s real-world applicability.

\subsubsection{Trajectory and Time-to-Collision Prediction Performance}

\begin{table}[b]
\centering
\renewcommand{\arraystretch}{1.5} 
\begin{tabular}{m{0.3cm} |cc}
\toprule
 & \textbf{Test Dataset 1} & \textbf{Test Dataset 2} \\
\midrule 

\centering \rotatebox{90}{Horizontal Pred.} & {\rowincludegraphics[width=0.42\linewidth]{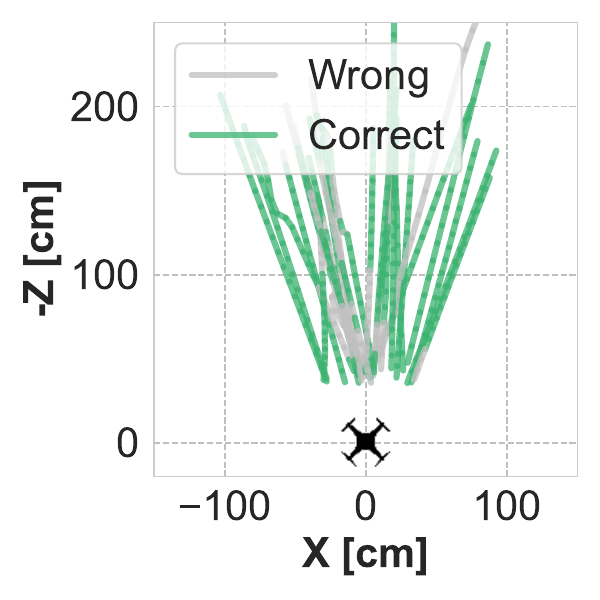}} &
{\rowincludegraphics[width=0.42\linewidth]{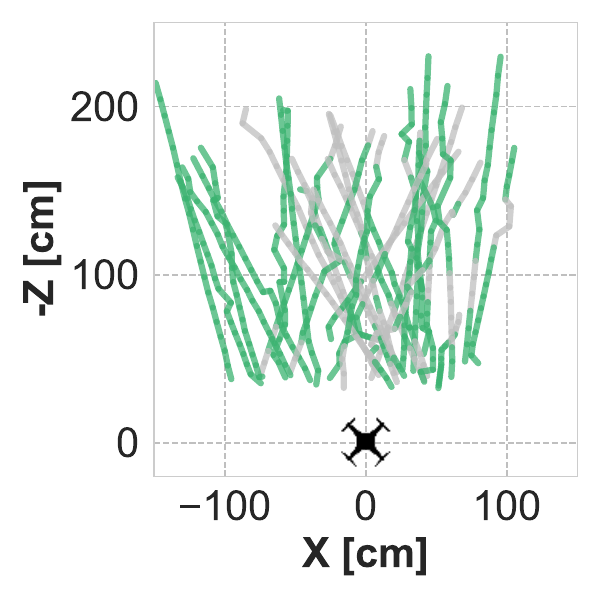}} \\ 

\centering \rotatebox{90}{Vertical Pred.} & {\rowincludegraphics[width=0.42\linewidth]{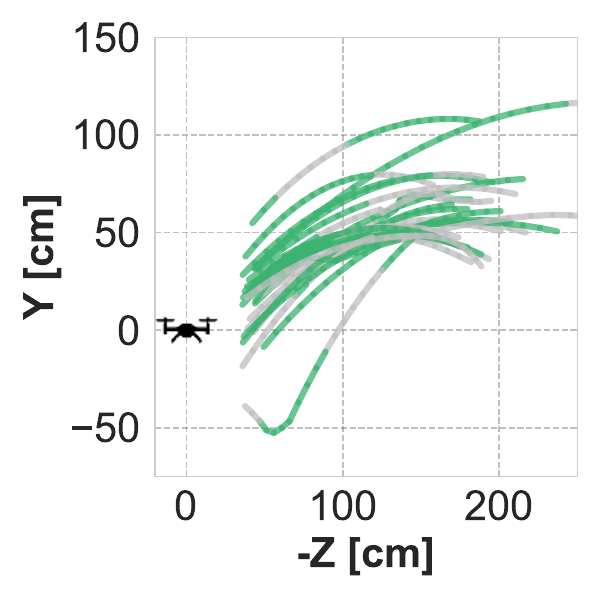}} &
{\rowincludegraphics[width=0.42\linewidth]{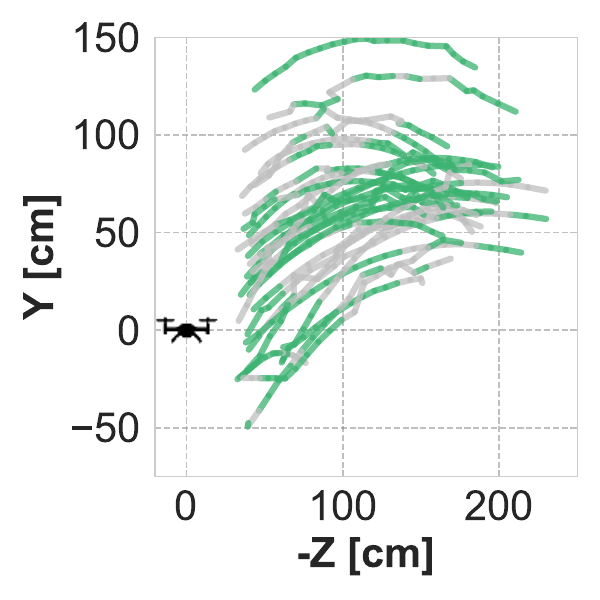}} \\ 

\centering \rotatebox{90}{Time Pred.} & {\rowincludegraphics[width=0.42\linewidth]{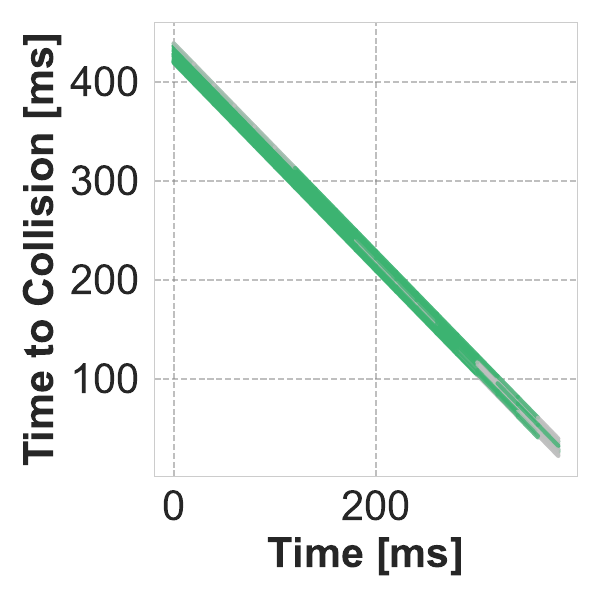}} &
{\rowincludegraphics[width=0.42\linewidth]{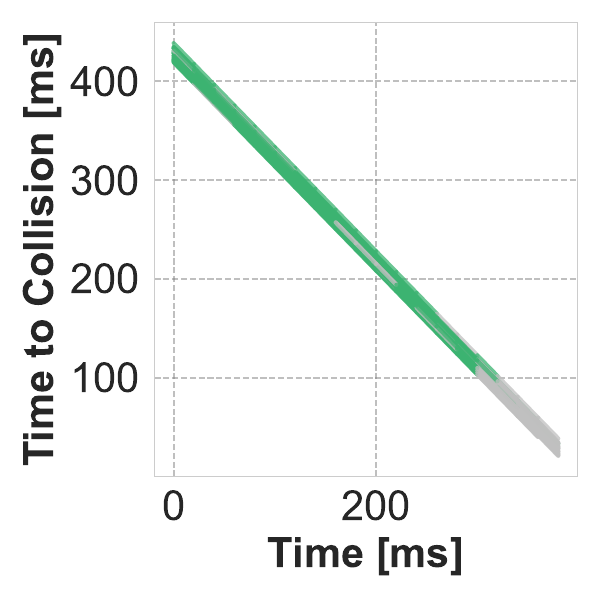}} \\ 

\bottomrule
\end{tabular}
    \caption{\textbf{Prediction performance plotted on the horizontal, vertical, and time-to-collision dimensions.} Green trajectories indicate correct predictions, while gray lines denote mispredictions.}
\label{fig:trajectory_performance}
\end{table}

In Table \ref{fig:trajectory_performance}, we illustrated the model prediction performance in horizontal, vertical and time-to-collision dimensions for all the recording individually across both test datasets. Green trajectories denote correct predictions, while gray indicates mispredictions.

In the horizontal and vertical prediction plots, Test Dataset 1 exhibits structured and converging trajectories. In contrast, Test Dataset 2 displays a more scattered distribution with strong variance. Notably, the mispredictions on both test datasets for left and right directions primarily occur in the center of the recordings, where the trajectories become less distinct, making it harder for the model to accurately predict the motion.

Overall, these results demonstrate the ability of the model to generalize well in the temporal and spatial domain, with a slight degradation in performance cause by higher variance in the second test dataset.

\subsection{Latency Results}

\begin{figure}[h]
    \centering
    \includegraphics[width=0.47\textwidth]{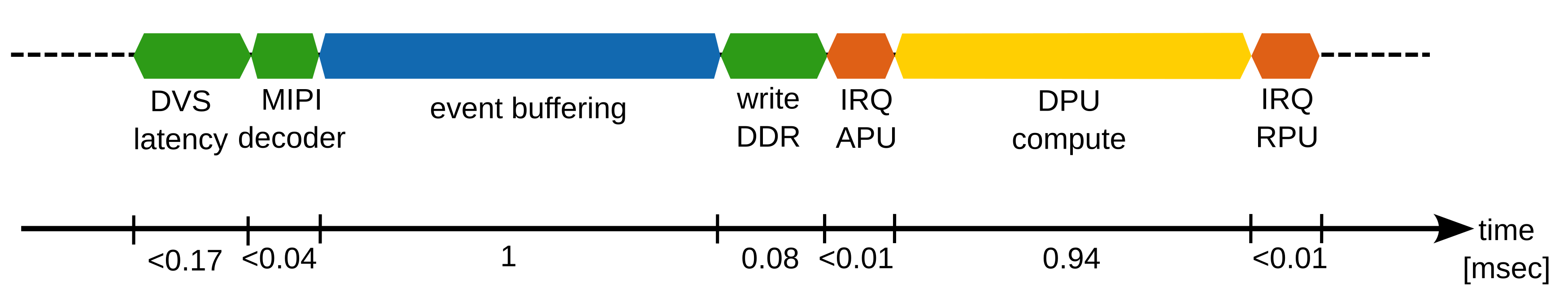}
    \caption{\textbf{End-to-end latency} split into the individual compute parts, including all data movements from EVS to final decision on the RPU.}
    \label{fig:hardware_latency}
\end{figure}

The hardware / software codesign implemented is shown in \Cref{fig:hardware_latency}, showcasing its worst-case end-to-end latency. The \ac{EVS} latency, as estimated in \cite{vogt2024IROS} and \cite{Gallego2022}, is approximately \qty{170}{\micro\second}. The decoding of the CSI2 MIPI data stream from \ac{EVS} takes less than \qty{40}{\micro\second}. Afterward, the events are decoded and aggregated with a variable integration time, which constitutes the largest latency component, with a value of \qty{1}{\milli\second} chosen for this step. Once aggregation is complete, writing to \ac{DDR} memory and signaling the APU requires approximately \qty{90}{\micro\second}. Inference on the FPGA accelerator runs within \qty{0.94}{\milli\second}. Finally, the RPU is signaled to read the inference results and, using a simple decision tree, determines the action to execute on the flight controller.

Compared to the EVS system, an RGB-based design would suffer from longer delay in data acquisition, caused by the frame-rate of the camera \qty{20}{\milli\second},. This disparity in processing speeds highlights the advantage of using event-based vision systems (EVS) for applications requiring low-latency decision-making.

\section{Future Work}
\label{sec:conclusion}

While this work demonstrates the effectiveness of EVS-based action prediction for UAV collision avoidance, several directions remain for future exploration. One avenue involves extending the model to handle more complex, multi-agent scenarios. Another promising direction is optimizing the FPGA deployment further by leveraging logic-gate networks \cite{petersen2022deepdifferentiablelogicgate} and exploring alternative hardware accelerators to push the limits of ultra-low-latency inference. 
Additionally, fusing event and RGB data, potentially in combination with stateful architectures such as recurrent or spiking neural networks, could enhance temporal understanding and improve robustness. Another important direction involves expanding the dataset with recordings captured during drone flight. Finally, future work could investigate self-supervised learning strategies to reduce the reliance on motion capture systems and improve the generalization in real-world applications.

\section{Conclusion}
\label{sec:conclusion}

The experimental results provide compelling evidence for the advantages of event-based vision systems over conventional RGB approaches in dynamic and latency-critical applications. Our analysis shows that EVS models not only achieve significantly lower temporal errors (approximately 20 ms improvement) and superior spatial prediction accuracy (approximately 30 mm improvement) at substantially higher effective frame rates (up to 1 kHz), but they also exhibit enhanced robustness in action classification.
Specifically, while RGB models benefit from temporal aggregation at lower frame rates, they still lag behind EVS models in both spatial and temporal precision. The EVS model consistently maintained lower errors in collision prediction and demonstrated a higher capacity to generalize across varying test conditions. achieving a 59 percentage point improvement in classification accuracy (78\% versus 19\%) and a significantly higher F1 score (0.73 versus 0.06). Furthermore, the overall processing pipeline operating within strict temporal constraints (under \qty{3}{\milli\second}) underscores the benefits of integrated hardware/software co-design. 
\newpage

{
    \small
    \bibliographystyle{ieeenat_fullname}
    \bibliography{main}
}


\end{document}